\documentclass[conference]{IEEEtran}
\IEEEoverridecommandlockouts
\usepackage{times}
\usepackage{epsfig}
\usepackage{graphicx}
\usepackage{subfigure}
\usepackage{multirow}
\usepackage{multicol}
\usepackage{cite}
\usepackage{amsmath,amssymb,amsfonts}

\usepackage{textcomp}
\usepackage{xcolor}

\usepackage{algpseudocode}  

\begin{document}

\title{SA-GD: Improved Gradient Descent Learning Strategy with Simulated Annealing }

\author{\IEEEauthorblockN{Zhicheng Cai 
}
\IEEEauthorblockA{\textit{School of Electronic Science and Engineering}\\
\textit{Nanjing University},
Nanjing, China 210023\\ Email: 181180002@smail.nju.edu.cn
}}

\maketitle

\begin{abstract}
   Gradient descent algorithm is the most utilized method when optimizing machine learning issues. However, there exists many local minimums and saddle points in the loss function, especially for high dimensional non-convex optimization problems like deep learning. Gradient descent may make loss function trapped in these local intervals which impedes further optimization, resulting in poor generalization ability. This paper proposes the SA-GD algorithm which introduces the thought of simulated annealing algorithm to gradient descent. SA-GD method offers model the ability of mounting hills in probability, tending to enable the model to jump out of these local areas and converge to a optimal state finally. We took CNN models as an example and tested the basic CNN models on various benchmark datasets. Compared to the baseline models with traditional gradient descent algorithm, models with SA-GD algorithm possess better generalization ability without sacrificing the efficiency and stability of model convergence. In addition, SA-GD can be utilized as an effective ensemble learning approach which improves the final performance significantly.
\end{abstract}

\section{Introduction}
It has been decades since the concept of machine learning was raised. From simple linear models to more complex issues processing high dimensional data such as reinforcement learning ~\cite{wiering2012reinforcement} and deep learning models ~\cite{goodfellow2016deep,lecun2015deep}, the machine learning community prospers continuously because of unremitting research. What’s more, machine learning has been successfully applied in many significant realms and tasks such as natural language processing ~\cite{wolf2020transformers}, computer vision ~\cite{voulodimos2018deep}, and speech recognition ~\cite{bhatt2021continuous}.  

Most of the machine learning models can be regarded as a kind of unconstrained optimization problem, requiring the optimal parameters to obtain better representation ability. Gradient descent algorithm ~\cite{ruder2016overview}, which is a kind of iterative method utilized to solve nonlinear equations, is generally applied to optimize machine learning issues such as linear regression ~\cite{gross2012linear}, multi-layer perceptron ~\cite{tolstikhin2021mlp}, convolutional neural networks ~\cite{he2016deep}, and so on ~\cite{vaswani2017attention}. Specifically, when facing a machine learning issue, we utilize gradient descent iteratively to minimize the well-designed target function and update these learned parameters by back-propagation ~\cite{li2012brief} to make model convergence. For tasks with tremendous dataset, stochastic gradient descent (SGD) algorithm ~\cite{ketkar2017stochastic} which splits dataset into mini-batches is utilized to balance between the speed and stability of model convergence. 

However, in the case of high dimensional non-convex optimization problems, there exists many intractable circumstances like local minimums, saddle points and plateaus in the loss function~\cite{goodfellow2016deep}. It is easy for optimization with gradient descent to make the model get into these disappointing areas and fail to jump out of inevitably. These situation tackles further optimization and learning. As a consequence, the model trained fails to obtain sufficient robustness and a good generalization ability. Therefore, many improved gradient descent algorithm have been raised such as NAG ~\cite{lin2019nesterov}, Adam ~\cite{kingma2014adam} and so on ~\cite{shi2020rmsprop}. However, these improved optimization algorithms still keep loss function performing gradient descent without possessing the ability of mounting hills ~\cite{goodfellow2016deep} and fail to solve certain issues mentioned above. As a matter of fact, researchers prefer utilizing stochastic gradient descent simply in many tasks ~\cite{krizhevsky2012imagenet,he2016deep}. In addition, these improved optimization algorithms fail to achieve better performance compared to simple SGD under some circumstances.

It is believed that quipping machine learning models with the ability of mounting hills is one approach to make the loss function jump out off these disappointing local areas. In addition, we believe that the ability of mounting hills should be according to certain probability function. Moreover, we argue this probability function should take all present model states into consideration, including loss, epoch, even learning rate. To take effect while ensuring model convergence, this probability should be low in the beginning and then ascending, but under a certain ceiling all the time. This paper set the upper bound as 33 percent. 

Inspired by the simulated annealing algorithm ~\cite{van1987simulated} whose probability function takes energy and temperature into consideration, we proposed SA-GD optimization algorithm which stands for gradient descent improved by simulated annealing. SA-GD algorithm possesses the gradient ascent probability function in the similar format as simulated annealing algorithm does. This function enables the model to perform gradient ascent in probability. With the ability of mounting hills, the model is supposed to jump out of these intractable local areas and converge to an optimal state. Moreover, the probability of performing gradient ascent can be regarded as a sort of noisy disturbance, which enhances the model generalization ability as well. In section 3 we will exhibit the derivation process of the final gradient ascent probability function step by step.

In addition, due to the fluctuation of loss curve caused by gradient ascent, the models of various epochs during training procedure are supposed to be different significantly. Make the precondition that all models selected perform well individually, the group of selected models possesses diversity. As a result, the thought of ensemble learning with SA-GD is proposed naturally.

\section{Related Work}

\subsection{Gradient descent algorithm}
For most deep learning and neural network algorithms, we utilize the gradient descent algorithm to optimize the minimization of the cost function. Suppose the current cost function is $J(\theta_i)$ and $J^{\prime}(\theta_i)$ is the derivative of the cost function at $\theta_i$, which is also the direction of the gradient that reduces the cost function the fastest. We can gradually reduce the value of the cost function by moving the parameter $\theta_i$ one small step in the derivative direction so that $\theta_{i+1} = \epsilon J^{\prime}(\theta_i)$ to obtain $J(\theta_{i+1}) < J(\theta_i)$, where $ \epsilon$ is the hyper-parameter learning rate. Iterating over the above process in the hope of eventually obtaining the global optimal solution is the gradient descent algorithm.

\begin{figure}[htbp]
\centering
\includegraphics[height=0.25\textwidth]{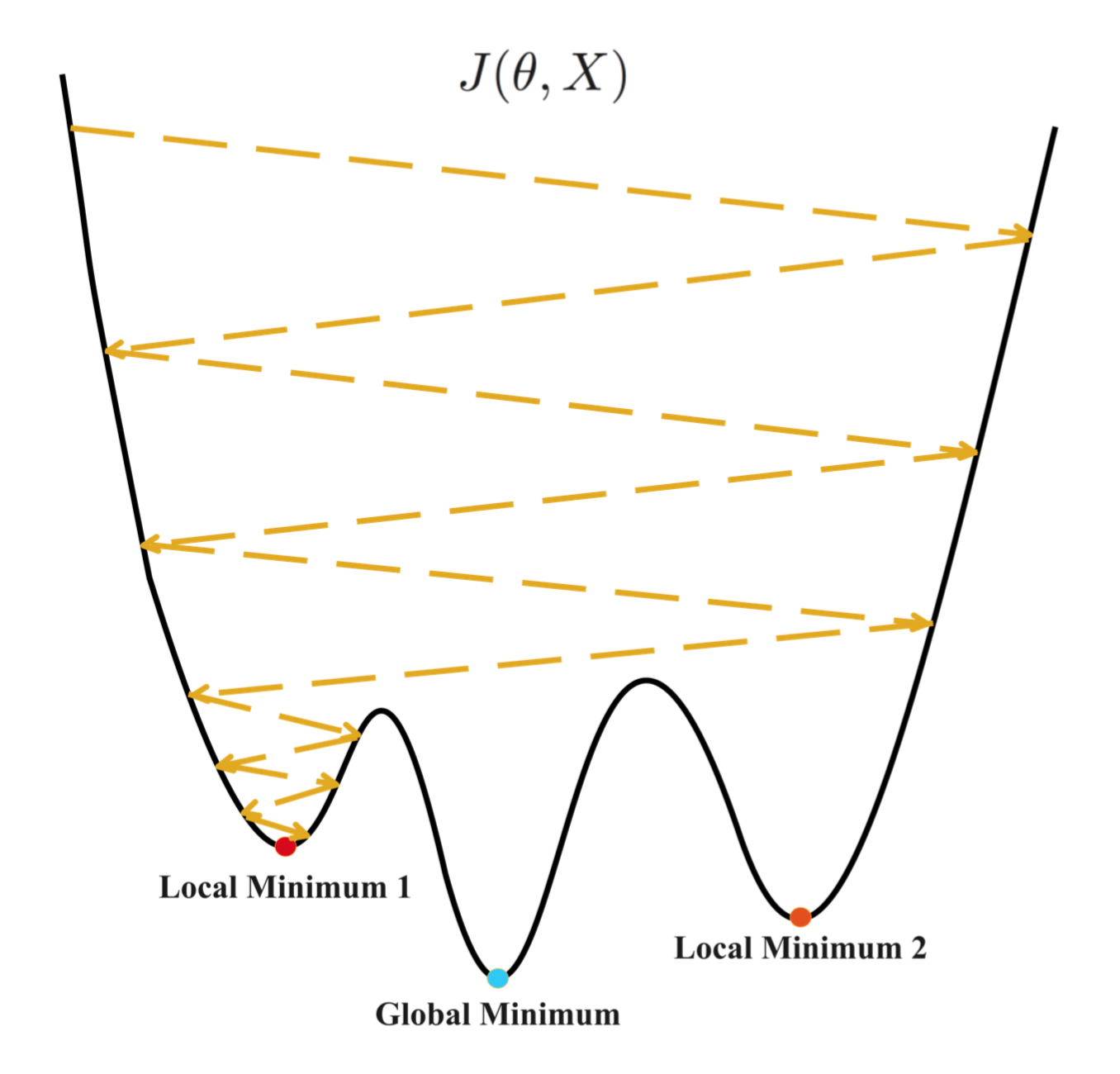}
\caption{Schematic diagram of the case where cost function falls into a local minimum.}
\label{p1}
\end{figure}

However, for a high-dimensional non-convex optimization problem like deep learning, there are often many saddle points, local minimums, or plateaus ~\cite{goodfellow2016deep}. Gradient descent may make the cost function fall into such local intervals and fail to jump out and converge to the global optimal solution. Fig.~\ref{p1} shows the case where the cost function falls into a local minimum. If the learned weights of the network are not initialized properly, it is very easy to make the network eventually converge to a poor state ~\cite{goodfellow2016deep}.

Stochastic gradient descent ~\cite{ketkar2017stochastic} takes one random sample at a time to calculate the gradient, and since the optimal direction of a single sample point may not be the global optimal direction, the possibility that the gradient is not zero when it falls into a local minimum is preserved, giving the loss function a chance to jump out of the local minimum. However, the stochastic approach can be seen as adding a random factor to the direction of the gradient and may provide the wrong direction of optimization, resulting in the network eventually failing to converge to a better state. For deep learning models, the probability of saddle point appearance is much higher than the probability of local minimum appearance ~\cite{goodfellow2016deep}, some improved gradient descent algorithms such as gradient descent with standard momentum term~\cite{qian1999momentum} or Nesterov momentum ~\cite{lin2019nesterov}, and adaptive learning rate optimization algorithms starting from the perspective of learning rate such as AdaGrad ~\cite{lydia2019adagrad} and AdaDelta ~\cite{zeiler2012adadelta}, trying to make the model escape from the saddle point and get a better convergence result. However, our experimental results validate that these improved optimization algorithms fail to make a significant improvement to the model generalization ability.

\subsection{Simulated annealing algorithm}
Researches in statistical mechanics have demonstrated that if a system is slowly cooled down (i.e. annealing process) starting from a high temperature, the energy of the system is steadily reduced and the state of the system particles is continuously changed. After sufficient transitions, thermal equilibrium can be reached at each temperature. When the system is completely cooled, it will enter a state of minimal energy with a high probability~\cite{du2016simulated}. The Metropolis algorithm ~\cite{beichl2000metropolis} constructs a mathematical model to simply describe the process of the system annealing. Assuming that the energy of the system in state $n$ is $E(n)$, the probability that the system transitions from state $n$ to state $n+1$ at temperature $T$ can be formulated as Eq.~\ref{eq1}.

\begin{equation}
   \begin{split}
            P(n\to n+1) =\left\{
\begin{aligned}
1 \ \ \ \ \ \ \ \  & , & E(n+1) \le E(n) \\
e^{\frac{E(n)-E(n+1)}{KT}}  & , & E(n+1) > E(n)
\end{aligned}
\right.
   \end{split}
   \label{eq1}
\end{equation}

where $K$ is the Boltzmann constant and $T$ is the system temperature.

Under a certain temperature, the system will reach thermal equilibrium after sufficient conversions have taken place. The probability of the system being in state $i$ at this point satisfies the Boltzmann distribution ~\cite{du2016simulated} as illustrated in Eq.~\ref{eq2}. 

\begin{equation}
P_T(X=i)=\frac{e^{-\frac{E(i)}{KT}}}{\sum_{j\in S}e^{-\frac{E(j)}{KT}}}
\label{eq2}
\end{equation}

where $S$ is the set of state spaces.

When the temperature drops very lowly to 0, there is:

\begin{equation}
   \begin{split}
   &\lim_{T\to 0}\frac{e^{-\frac{E(i)-E_{min}}{KT}}}{\sum_{j\in S}e^{-\frac{E(j)-E_{min}}{KT}}} \\
   &=\lim_{T\to 0}\frac{e^{-\frac{E(i)-E_{min}}{KT}}}{\sum_{j\in S_{min}}e^{-\frac{E(j)-E_{min}}{KT}}+{\sum_{j\notin S_{min}}e^{-\frac{E(j)-E_{min}}{KT}}}} \\
            &=\lim_{T\to 0}\frac{e^{-\frac{E(i)-E_{min}}{KT}}}{\sum_{j\in S_{min}}e^{-\frac{E(j)-E_{min}}{KT}}} \\
            &=\left\{
\begin{aligned}
\frac{1}{|S_{min}|} \ \  & , & i\in S_{min} \\
0 \ \ \ \ \ \  & , & otherwise
\end{aligned}
\right.
   \end{split}
   \label{eq3}
\end{equation}

where $E_{min}$ represents the lowest energy in the state space and $S_{min}={i|E(i)=E_{min}}$ stands for the set of states with the lowest energy. Eq.\ref{eq3} illustrates that when the temperature drops to a very low level, the system will enter the minimal energy state with a large probability.

Inspired by statistical mechanics, the idea of simulated annealing is utilized for the optimization problem of searching for the global minimum. Consequently we obtain the simulated annealing algorithm ~\cite{du2016simulated}. Suppose the evaluation function of the problem is $f(x)$, the temperature at a certain moment is $T_i$, one feasible solution of the problem is $x_t$, and the solution that can be generated is $x^{\prime}$. Afterwards, the probability that the new solution $x^{\prime}$ is accepted as the next new solution $x_{t+1}$ is:

\begin{equation}
   \begin{split}
            P(x_n\to x' ) =\left\{
\begin{aligned}
1 \ \ \ \ \ \ \ \  & , & f(x') \le f(x_n) \\
e^{-\frac{f(x')-f(x_n)}{T_i}}  & , & f(x') > f(x_n)
\end{aligned}
\right.
   \end{split}
   \label{eq4}
\end{equation}

At the temperature $T_i$, after several state transitions, the temperature is reduced to $T_{i+1} < T_i$. The processes are repeated again at temperature $T_{i+1}$. Since each new state only completely depends on the previous one, this is a classic Markov process. As a consequence, the probability of generating the solution $x^{\prime}$ from $x_t$ is uniform. After finitely many transitions, the distribution of the equilibrium state $x_t$ at temperature $T_i$ is:

\begin{equation}
P_t(T_i)=\frac{e^{-\frac{f(x_t)}{T_i}}}{\sum_{j\in S}e^{-\frac{f(x_j)}{T_i}}}
\label{eq5}
\end{equation}

When the temperature drops to 0, there is:

\begin{equation}
   \begin{split}
   P_i^*(T=0)=\left\{
\begin{aligned}
\frac{1}{|S_{min}|} \ \  & , & x_i^*\in S_{min} \\
0 \ \ \ \ \ \  & , & otherwise
\end{aligned}
\right.
   \end{split}
   \label{eq6}
\end{equation}

If the temperature decreases very slowly, in addition, a sufficient number of state transitions are made at each temperature, the system can reach thermal equilibrium at each temperature. As a result, the global optimal solution will be found with probability 1.

Boltzmann machine ~\cite{hinton2007boltzmann}, which is a kind of stochastic neural network, borrows the idea of simulated annealing. Instead of adjusting the weight parameters based on deterministic algorithms like other networks, the BM network modifies the weights with some probability distribution during the training stage. In the inference stage, the BM network also determines the transition of its states according to certain probability distribution. As the BM network states evolving, the energy of the network always changes in the direction of decreasing in the probabilistic sense.  However, some neuron states may temporarily increase the network energy due to small probability taking values. This gives the BM network the ability to "mount the hill" by jumping out of the local minimum. Another method which applies simulated annealing algorithm to artificial neural networks ~\cite{sexton1999beyond} obtains the weights of candidate network neurons by Monte Carlo sampling and generating function. Then simulated annealing algorithm is utilized to judge whether selecting the candidate weights as new weights. The process is repeated continuously to make the neural network converge to a better state. However, it is usually very time consuming to search for the optimal parameters utilizing this method. In the SA-GD algorithm, the probability function of simulated annealing is used as the probability of gradient ascent for the model cost function.

\section{Simulated Annealing Gradient Descent}

\subsection{Utilizing the thought of simulated annealing to aid training}
First we make initial improvements to the gradient descent algorithm by drawing on the idea of simulated annealing. As an example, this paper utilizes the universal mini-batch stochastic gradient descent algorithm ~\cite{khirirat2017mini} to train the convolutional neural network to complete the image classification task. The objective function of this task is the cross-entropy loss function $J(\theta)$, which represents the energy of the current system. For the $i$-th, $i \in [1,2,3,......] $ mini-batch stochastic gradient descent, we first calculate the loss value $J(\theta_i)$ caused by the current weight parameters $\theta_i$. Currently, the energy value of the system is $E_i$ and the gradient is $\nabla_{\theta_i}{J(\theta_i)}$. Then compare it with the energy value $E_{i-1}$ of the previous state (the initial energy of the system is set to be $E_0=0$) and calculate the difference $\triangle E_i $ between them. Afterwards, the transition probability $P_i$ is set as:

\begin{equation}
P_i = e^{-\frac{|\triangle E|}{T_0\cdot \ln(n+1)}}
\label{eq7}
\end{equation}

where $n, n\in [1,2,3,......] $ is the number of the current iteration epochs, the current temperature is $T=-T_0 \cdot \ln(1)$. We update the parameters by accepting the $i$-th gradient descent with probability $1-P_i$ and perform the reverse gradient ascent with probability $P_i$ to update the parameters. Suppose the current learning rate is $\epsilon$, it can be stated in mathematical language as below:

Considering the probability $P_i$ as the threshold for performing the transition, and obtaining $a$ by random sampling from a uniform distribution with the interval [0,1], the expression of the weight parameter at the next moment is:

\begin{equation}
   \begin{split}
            \theta_{i+1} =\left\{
\begin{aligned}
 \theta_i - \epsilon \cdot \nabla_{\theta_i}{J(\theta_i)} & , & a < P_i \\
  \theta_i + \epsilon \cdot \nabla_{\theta_i}{J(\theta_i)}& , & a  \ge P_i
\end{aligned}
\right.
   \end{split}
   \label{eq8}
\end{equation}

In order to adjust the size of the fluctuations, namely, the capacity of the mounting the hill, we can add the corresponding hyper-parameter expansion factor $\sigma$  to Eq.~\ref{eq8}. At this point, the weight update expression becomes:

\begin{equation}
   \begin{split}
            \theta_{i+1} =\left\{
\begin{aligned}
 \theta_i - \epsilon \cdot \nabla_{\theta_i}{J(\theta_i)} & , & a < P_i \\
  \theta_i + \sigma \cdot \epsilon \cdot \nabla_{\theta_i}{J(\theta_i)}& , & a  \ge P_i
\end{aligned}
\right.
   \end{split}
   \label{eq9}
\end{equation}

In this way, we want to endow the model with a probabilistic hill-mounting capability, which enables small fluctuations while making the loss function show a downward trend in general, so that the loss function can jump out of the local minimum or saddle point region and finally converge to a better state. At the same time, such gradient ascent is equivalent to adding noise to the training process, which plays a slight regularization effect and improves the generalization ability of the model. Subsequent experimental results validate that such a preliminary improved simulated annealing-assisted gradient descent algorithm can bring significant test accuracy improvement on the CIFAR-10 benchmark dataset.

\subsection{Algorithm description}
This section provides the corresponding additional description of the above algorithm.

On the one hand, in the above algorithm, we do not perform state probability transition in the same way as the simulated annealing algorithm. If the idea of simulated annealing is followed exactly, for the $i$-th mini-batch gradient descent, the current parameter $\theta_i$ causes a loss value of $J(\theta_i)$, that is, the energy value of the system is $E_i$ and the gradient is $\nabla_{\theta_i}{J(\theta_i)}$. Firstly, perform gradient descent to obtain the candidate weight value $\theta_i ^{\prime}=\theta_i - \epsilon \nabla_{\theta_i}{J(\theta_i)} $ and calculate the loss value $J(\theta_i ^{\ prime})$. Consequently, the energy of the system after performing gradient descent is $E_i ^{\prime}$. At this point, the probability of whether to accept $\theta_i '$ as a new solution $\theta_{i+1} = \theta_i - \epsilon \cdot \nabla_{\theta_i}{J(\theta_i)}$ is:

\begin{equation}
   \begin{split}
            P(\theta_i\to \theta' ) =\left\{
\begin{aligned}
1 \ \ \ \ \ \ \ \  & , & E(\theta') \le E(\theta_i) \\
e^{-\frac{E(\theta' )-E(\theta_i )}{T_0 \cdot \ln(n+1)}}  & , & E(\theta' )> E(\theta_i )
\end{aligned}
\right.
   \end{split}
   \label{eq10}
\end{equation}

If $E(\theta^{\prime}) > E(\theta_i)$ and $\theta_i ^{\prime}$ is not accepted, since the candidate parameter values are deterministic, the parameter values remain unchanged at this point and fail to be updated. Moreover, if $E(\theta^{\prime}) \le E(\theta_i)$ or $E(\theta^{\prime}) > E(\theta_i)$ and $\theta_i ^{\prime}$ is accepted with probability, then the algorithm is indistinguishable from an ordinary gradient descent algorithm except that the convergence rate is much slower. One possible improvement is to make the use of a random gradient, which makes the candidate weights random rather than definitized. However, such an approach will lead to optimization in the wrong direction or even fails to converge better.

On the other hand, instead of using the difference directly, we use the absolute value of the difference. This design enables a probability of gradient ascent regardless of whether the weights obtained from the next gradient update make the loss increase or decrease, giving the model the ability to mount hills. Since the current direction of gradient descent is not necessarily the direction of global minimum, if we accept all the weight parameters for making the loss reduction, we do not make an improvement on whether we can escape the local optimum or saddle point. As a result, we should instead give the model some hill mounting ability when the loss is reduced. In addition, there may also be structures with large gradients like cliffs in deep learning optimization. When the parameters approach such a cliff region, gradient descent updates can make the parameters vary very much, ejecting the loss function very far away and possibly making a lot of previous optimization work meaningless ~\cite{goodfellow2016deep}. However, if the gradient ascends at this point, it can make the loss function have a better chance to enter the bottom of the cliff at the next gradient descent, avoiding the huge span while getting a good optimization in this way.

\begin{figure}[htbp]
\centering
\includegraphics[height=0.25\textwidth]{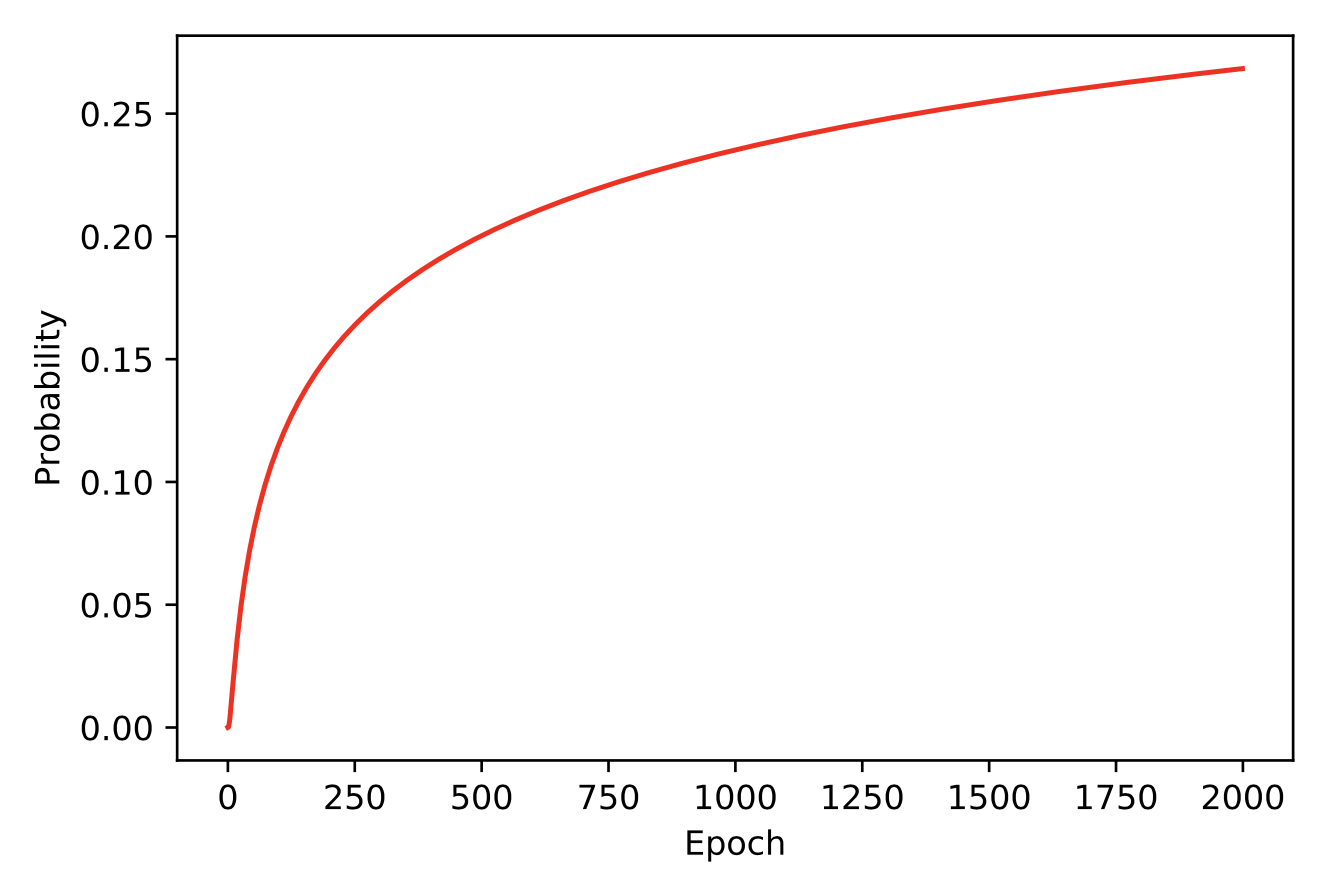}
\caption{Schematic diagram of the transition probability change curve without considering the learning rate.}
\label{p2}
\end{figure}

Moreover, in order to ensure model convergence, the probability of performing gradient ascent needs to be absolutely less than 50\% (the transition probability is set to be less than 33\% in this paper). In addition, this probability should increase with the number of iteration epochs, which is referred to as a prerequisite for using simulated annealing assisted training in this paper. That is, in the first few epochs of training, the cost function descends with a large probability (90\%). In the following epochs, the cost function ascends with a gradually increasing but constant probability of less than 33\%. In each epoch, the annealing temperature is kept constant. As a consequence, for each mini-batch of gradient descent the candidate values are calculated, the states can be fully transited at that epoch temperature. When an epoch is trained completely, annealing is performed by the formula $T=-T_0 \cdot \ln(n+1)$. Through extensive experiments, it was shown that the training loss of CIFAR-10 eventually converged by an order of magnitude of $0.0001$. Therefore, in the subsequent related experiments we set the initial temperature $T_0=0.0001$ and utilized a constant learning rate $\epsilon=0.001$. The transition probability change curve obtained currently is shown in Fig.~\ref{p2}.

\subsection{Considering the learning rate}

In the above preliminary improvement, we did not take into account the role of the learning rate in the expression of the transition probability. Now we add the hyper-parameter learning rate to the transfer probability expression, at which point the transfer probability $P_i$ is:

\begin{equation}
P_i = e^{-\frac{|\triangle E|}{T_0 \cdot \epsilon \cdot \ln(n+1)}}
\label{eq11}
\end{equation}

\begin{figure}[htbp]
\centering
\includegraphics[height=0.25\textwidth]{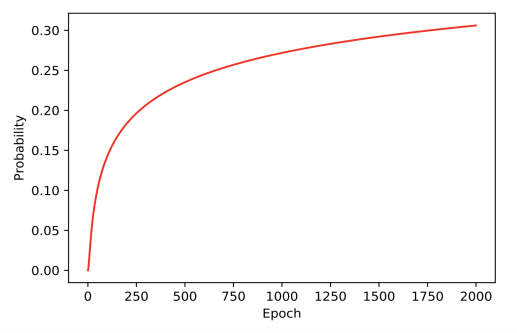}
\caption{Schematic diagram of the transition probability change curve considering the fixed learning rate.}
\label{p3}
\end{figure}

We first consider the simple fixed learning rate strategy. Set the learning rate to be constant at $\epsilon=0.001$. In order to meet the precondition, the initial temperature $T_0$ needs to have an order of magnitude $0.1$. We get the better $T_0=\frac{1}{9}$ by exhaustive search, at which time the transition probability change curve is shown in Fig.~\ref{p3}.

\subsection{Double simulated annealing}

During the actual training of the model, the learning rate usually decays gradually. On the one hand, a larger learning rate in the initial phase of learning speeds up learning, while a smaller learning rate afterwards allows the model to converge better and helps to find a more optimal local solution ~\cite{goodfellow2016deep}. On the other hand, as the number of training epochs increases, the gradient pathology problem occurs, causing learning to become very slow despite the large gradient. As a consequence, the learning rate must be contracted to compensate for the stronger curvature ~\cite{goodfellow2016deep}.

\begin{figure}[htbp]
\centering
\includegraphics[height=0.25\textwidth]{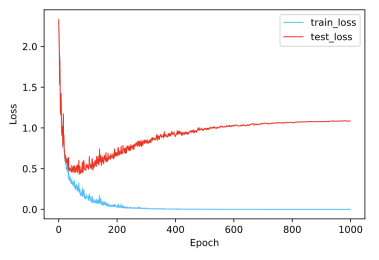}
\caption{The training and test loss curves obtained on CIFAR-10 dataset.}
\label{p4}
\end{figure}

\begin{figure}[htbp]
\begin{center}
\subfigure[Initial temperature $T_0=10$]{
\includegraphics[width=0.22\textwidth]{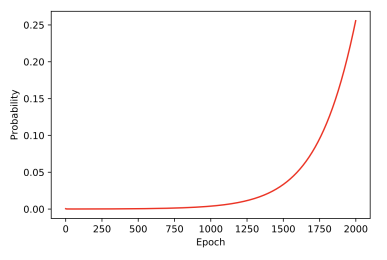}}
\subfigure[Initial temperature $T_0=0.1$]{
\includegraphics[width=0.22\textwidth]{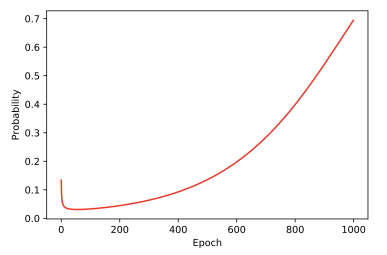}}
\end{center}
\caption{Schematic diagram of transition probability curves with different initial temperatures}
\label{p5}
\end{figure}

The learning rate strategy utilized in the subsequent experiments of this paper is an exponentially decay learning rate strategy with an initial learning rate set to 0.1 and a hyper-parameter $\gamma=0.9956$. If the transition probability is still obtained according to the Eq.~\ref{eq11}, it is difficult to obtain a suitable transition probability, although the initial temperature $T_0$ can be selected to satisfy the precondition. If the selected initial temperature $T_0$ is large, the transition probability will be very small in the first half of the training period and suddenly increase rapidly in the subsequent training epochs. Although such a changing transition probability can accelerate the model convergence, it is too small in the early stage and the model has almost no hill-mounting ability. However, the model using the exponentially decay learning rate strategy has almost converged at this time and cannot make better progress in generalization performance . If the selected initial temperature $T_0$ is small, it will make the moment of rapid increase of transition probability forward, resulting in a very high probability of performing gradient ascent when the model is approaching convergence, making it difficult to converge. This dilemma illustrates that a better transition probability cannot be obtained by simply adjusting the initial temperature. Fig.~\ref{p4} shows the training and test loss curves obtained on CIFAR-10 dataset. The transition probability curves obtained by picking a larger initial temperature $T_0=10$ and a smaller initial temperature $T_0=0.1$ (default $\triangle E = 0.001$) are shown in the Fig.~\ref{p5}.

From the training loss curve, the model converges quickly due to the exponentially decay learning rate strategy using a larger learning rate at the beginning, resulting in a rapid change in the loss difference. As shown by a large amount of experimental data, the loss difference $\triangle E $ decays by a factor of 10,000 to 100,000 within 200 epochs. In addition, the value of $\triangle E $ fluctuates significantly. Moreover, the learning rate of exponential decay decreases very fast, and these two factors make the transition probability difficult to control. Therefore, we first add a fractional power constraint to $\triangle E $ in the hope of attenuating the effect of the unstable fluctuations of $\triangle E $. Thus the transition probability $P_i$ becomes:

\begin{equation}
P_i = e^{-\frac{{|\triangle E|}^{\frac{1}{2}}}{T_0 \cdot \epsilon \cdot \ln(n+1)}}
\label{eq12}
\end{equation}

\begin{figure}[htbp]
\begin{center}
\subfigure[Curve of the fractional powers]{
\includegraphics[width=0.22\textwidth]{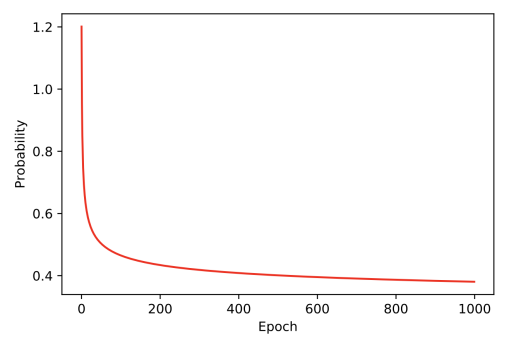}}
\subfigure[Curve of the temperature]{
\includegraphics[width=0.22\textwidth]{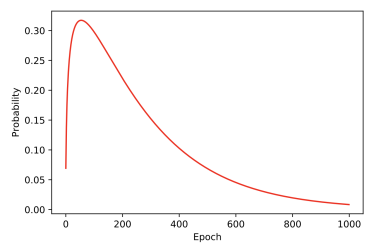}}
\end{center}
\caption{The variation curves of the fractional powers and the temperature}
\label{p6}
\end{figure}

\begin{figure*}[htbp]
\begin{center}
\subfigure[$\triangle E =0.001$]{
\includegraphics[width=0.30\textwidth]{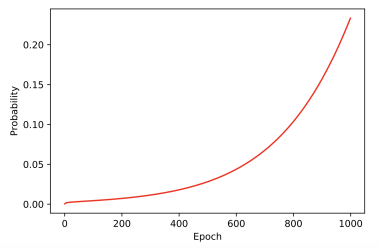}}
\subfigure[$\triangle E =0.0001$]{
\includegraphics[width=0.30\textwidth]{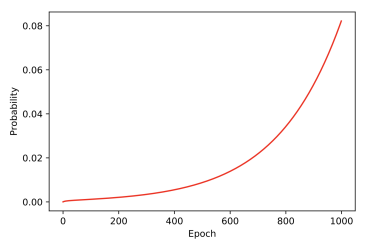}}
\subfigure[$\triangle E =0.00001$]{
\includegraphics[width=0.30\textwidth]{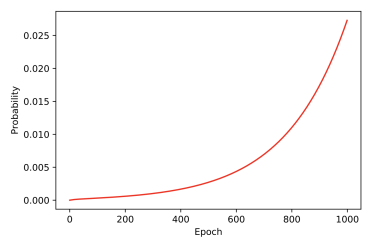}}
\end{center}
\caption{The curves of the transition probability with different $\triangle E$}
\label{p7}
\end{figure*}

\begin{figure*}[htbp]
\begin{center}
\subfigure[$\triangle E =0.001$]{
\includegraphics[width=0.30\textwidth]{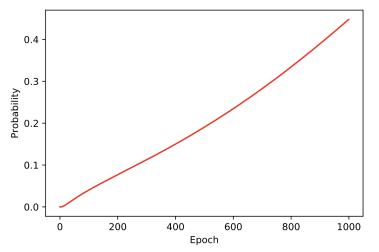}}
\subfigure[$\triangle E =0.0001$]{
\includegraphics[width=0.30\textwidth]{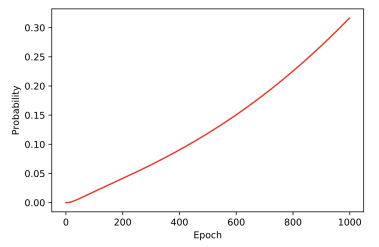}}
\subfigure[$\triangle E =0.00001$]{
\includegraphics[width=0.30\textwidth]{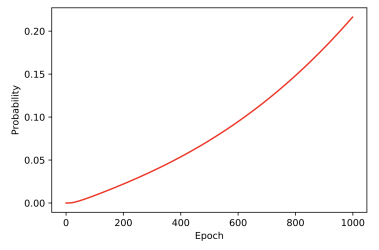}}
\end{center}
\caption{The curves of the transition probability constraining the fluctuation of temperature with different $\triangle E$}
\label{p8}
\end{figure*}

However, $\triangle E $ shows a huge decreasing trend, and the effect of the huge exponential decay of $\triangle E $ cannot be well bounded using a fixed exponential constraint. Therefore, we consider a fractional power constraint that can decrease as the number of training epochs increases. After performing sufficient experiments, we finally choose the fractional powers as ${\ln(n+2)}^{-\frac{1}{\alpha}}$ and choose the hyper-parameter $\alpha=e$. Fig.~\ref{p6} shows the variation curves of the fractional powers and the temperature $T = T_0 \cdot \epsilon_0 \cdot \gamma^n \cdot \ln(n+1)$. 

At this point the transfer probability $P_i$ becomes:

\begin{equation}
P_i = e^{-\frac{{|\triangle E|}^{{\ln(n+2)}^{-\frac{1}{\alpha}}}}{T_0 \cdot \epsilon \cdot \ln(n+1)}}
\label{eq13}
\end{equation}

Now the obtained transition probability curves with different $\triangle E$ ($T_0=15 \triangle E =0.001$,$\triangle E =0.0001$,$\triangle E =0.00001$) are shown in Fig.~\ref{p7}.

However, when the initial temperature is large, the transition probability will be very small, thus the ability of the model to climb the hill before convergence has been weak. Besides, when the initial temperature is small, the transition probability is therefore also large because the difference is large at the beginning of training, which is not conducive to model training and convergence. At this point, the fluctuating variation of temperature $T = T_0 \cdot \epsilon_0 \cdot \gamma^n \cdot \ln(n+1)$ becomes the primary reason for the difficulty in obtaining an optimal simulated annealing transition probability.Therefore, we choose to continue using fractional powers to reduce the impact of such fluctuations. At this point the transfer probability $P_i$ becomes:

\begin{equation}
P_i = e^{-\left({\frac{|\triangle E|^{\ln(n+2)^{-\frac{1}{\alpha}}}}{T_0 \cdot \epsilon_0 \cdot    \gamma^n \cdot \ln(n+2)}}\right)^{\frac{\beta}{\ln(n+2)}}}
\label{eq14}
\end{equation}

Where hyper-parameter $\beta$ is set to be the Euler–Mascheroni constant $0.5772$. The obtained transition probability curves with different $\triangle E$ ($T_0=15, \triangle E =0.001$, $\triangle E =0.0001$, $\triangle E =0.00001$) are shown in the Fig.~\ref{p8}. As shown in the Fig.~\ref{p8}, the transition probabilities so obtained are good at different magnitudes of $\triangle E$, endowing the model with favorable and suitable hill mounting ability. Since the learning rate is decaying, it can be regarded as an annealing process as well. Besides, the decaying learning rate is employed in the annealed transition probability, which is equivalent to performing a double annealing. Such heuristic improved gradient descent training algorithm can bring 0.7\% improvement to the model on the CIFAR-10 benchmark dataset.

\section{Experiments and Analysis}
To validate the effectiveness of SA-GD algorithm, we take convolutional neural network as a convincing instance. We compared our SA-GD algorithm with traditional gradient descent on various challenging benchmark datasets, including CIFAR-10 and CIFAR-100.

\subsection{Experimental configurations}

In this paper, the classic convolutional neural network model  VGG-16\cite{simonyan2014very} is selected as the basic model. VGG-16 is composed of 13 $3\times 3$ convolutional layers and 3 fully connected layers. Instead of employing pooling method, the convolutional layers with step-size is utilized to conduct down-sampling operation~\cite{cai2021study}. As a regularization method, we add the dropout ~\cite{srivastava2014dropout} to the first two of the three fully connected layers and set the activation rate to $50\%$. In addition, we add batch normalization layers ~\cite{ioffe2015batch} after all convolutional layers. The Xavier initialization strategy ~\cite{2010Understanding} is employed to initiate all the learned parameters. All the models utilize cross-entropy loss as the loss function. The SA-GD training algorithm adopts mini-batch with a momentum term of 0.95 and weight decay coefficient of 0.0005, the batch size is set to be 4096. As stated above, the learning rate strategies utilized including fixed learning rate $0.001$ and exponentially decay learning rate strategy with an initial learning rate set to 0.1 and the hyper-parameter $\gamma=0.9956$. The training epochs are set to be 2000 and 1000 respectively according to above two different learning strategies. All experimental models have been trained with NVIDIA GeForce GTX 2080Ti GPU.

\subsection{Dataset}

\textbf{CIFAR-10}: CIFAR-10 dataset is consisted of 10 classes of natural images with 50,000 training images and 10,000 testing images in total. The 10 classes include: airplane, automobile, bird, cat, deer, dog, frog, horse, ship and truck. Each image is a RGB image of size $32\times32$. 

\textbf{CIFAR-100}: CIFAR-100 dataset is consisted of 100 different classifications, and each classification includes 600 different color images, of which 500 are training images and 100 are test images. As a matter of fact, these 100 classes are composed of 20 super classes, and each super class possesses 5 child classes. The images in the CIFAR-100 dataset have a size of $32\times32$ like CIFAR-10. 

\textbf{SVHN}: SVHN (Street View House Numbers) dataset is composed of 630,420 RGB digital images with a size of $32\times32$, including a training set with 73,257 images and a test set with 26,032 images. 

\textbf{Fashion-MNIST}: Fashion-MNIST (FMNIST for short) is a dataset of 10 class fashion items, including T-shirt/top, trouser, pullover, dress, coat, scandal, shirt, sneaker, bag and ankle boot. Each sample is a gray image with a size of $28\times28$. This dataset contains 60,000 training images and 10,000 test images in total.

\textbf{Data augmentation}: For the training data of CIFAR-10 and CIFAR-100 data sets, we adopt the method of data augmentation as follows: four circles of zero pixels are padded around the original image, and then the padded image was randomly cropped to the size of the original image. Then we flip the image horizontally at a probability level of 0.5. 

\subsection{Compared approaches}
Six groups of comparative experiments are carried out on all the selected four data sets. There are:
\begin{itemize}
\item GD-Fixed stands for the methods employing traditional gradient descent algorithm with a fixed learning rate $0.001$ as the training strategy.
\item SA-Fixed represents the SA-GD training algorithm without considering the learning rate as depicted above, the initial temperature is set to be $T_0=0.0001$ and the learning rate $0.001$ is fixed.
\item SA-Fixed-LR stands for the SA-GD algorithm considering fixed learning rate $0.001$, the temperature is set to be $T_0=\frac{1}{9}$ initially. 
\item GD-Exp utilizing traditional gradient descent algorithm with exponentially decay learning rate strategy.
\item SA-Exp is the SA-GD algorithm without considering the learning rate, the initial temperature is set to be $T_0=0.0001$.
\item DSA-Exp stands for the SA-GD algorithm considering the exponentially decay learning rate as stated above, the initial temperature is set to be $T_0=15$.
\end{itemize}
To be seen more clearly, the configurations of these six compared methods are shown in Table.~\ref{t1}. Here, the hyper-parameter expansion factor $\sigma$ stated in Eq.~\ref{eq9} is set to be constant one. Moreover, in spite of the utilization of the fixed learning rate, the difference between method SA-Fixed and SA-Fixed-LR actually lies on the different choices of the initial temperature $T_0$.

\begin{table}[ht]
   \caption{Configurations of six compared methods.}
    \label{t1}
   \centering
   \scalebox{0.95}{
    \begin{tabular}{|c|c|c|c|c|}
    \hline
    Method & SA & LR strategy & Consider LR & $T_0$  \\
    \hline
    GD-Fixed & False  & Fixed & - & -  \\
    \hline
    SA-Fixed & True  & Fixed & False & 0.0001  \\
    \hline
    SA-Fixed-LR & True  & Fixed & True & $\frac{1}{9}$  \\
    \hline
    GD-Exp & False  & Exponentially decay & - & -  \\
    \hline
    SA-Exp & True  & Exponentially decay & False & 0.0001  \\
    \hline
    DSA-Exp & True  & Exponentially decay & True & 15  \\
    \hline
    \end{tabular}
    }
 \end{table}
 
\subsection{Results}
We tested all the methods for five times respectively on four challenging benchmark datasets, Table.~\ref{t2} exhibits the mean value of each five repeated experiments.  

\begin{table}[ht]
   \caption{Experimental results of six compared methods.}
    \label{t2}
   \centering
   \scalebox{0.95}{
    \begin{tabular}{|c|c|c|c|c|}
    \hline
    Method & CIFAR-10 & CIFAR-100 & SVHN & F-MNIST  \\
    \hline
    GD-Fixed & 88.83\% & 65.22\% & 92.73\% &  92.45\% \\
    \hline
    SA-Fixed & 89.56\% & 66.02\% & 93.24\% &  92.77\% \\
    \hline
    SA-Fixed-LR & 89.71\% & 65.98\% & 93.29\% & 92.73\%  \\
    \hline
    GD-Exp & 89.54\% & 67.21\% & 93.46\% &  92.71\% \\
    \hline
    SA-Exp & 90.17 & 68.05\% & 93.94\% &  92.94\% \\
    \hline
    DSA-Exp & \textbf{90.21\%} & \textbf{68.07\%} & \textbf{94.03\%} & \textbf{93.01}  \\
    \hline
    \end{tabular}
    }
 \end{table}
 
 \begin{figure*}[htbp]
\begin{center}
\subfigure[GD-Exp]{
\includegraphics[width=0.30\textwidth]{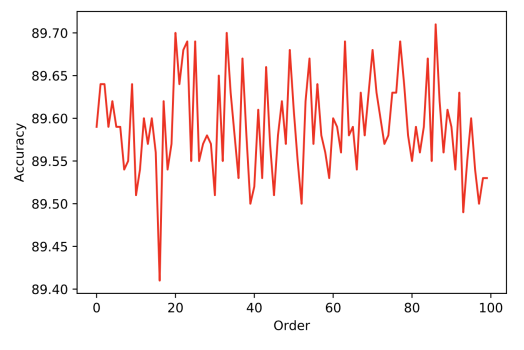}}
\subfigure[SA-Exp]{
\includegraphics[width=0.30\textwidth]{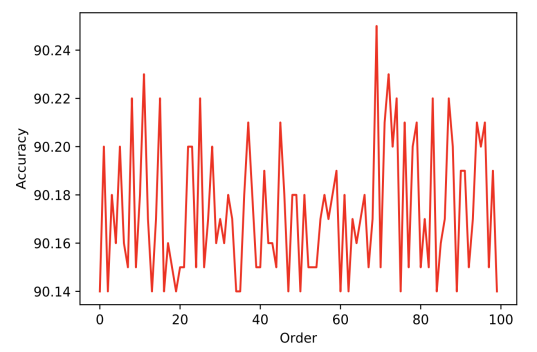}}
\subfigure[DSA-Exp]{
\includegraphics[width=0.30\textwidth]{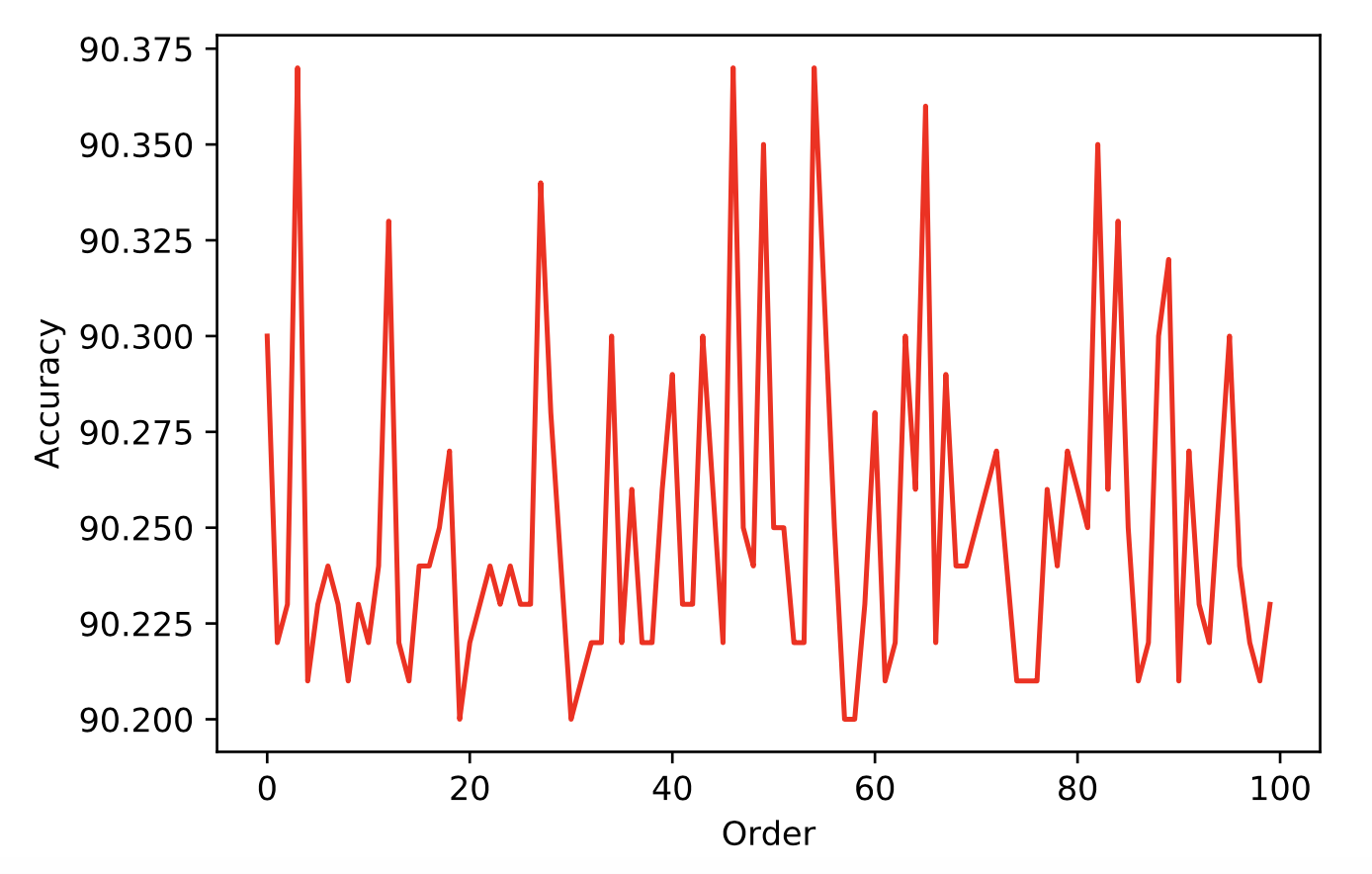}}
\end{center}
\caption{The experimental results of testing methods GD-Exp, SA-Exp, and DSA-Exp for one hundred times}
\label{p9}
\end{figure*}

\begin{figure*}[htbp]
\begin{center}
\subfigure[GD-Exp]{
\includegraphics[width=0.30\textwidth]{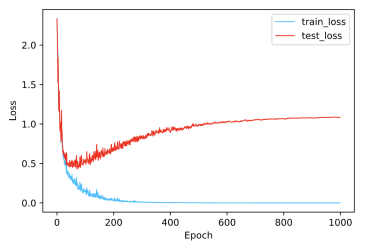}}
\subfigure[SA-Exp]{
\includegraphics[width=0.30\textwidth]{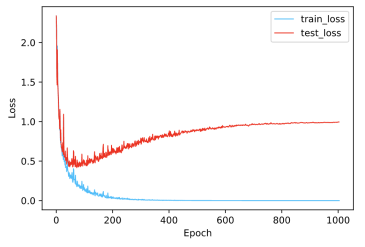}}
\subfigure[DSA-Exp]{
\includegraphics[width=0.30\textwidth]{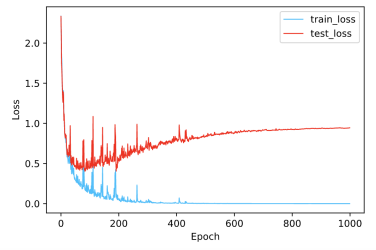}}
\end{center}
\caption{The training and test loss curves of methods GD-Exp, SA-Exp, and DSA-Exp}
\label{p10}
\end{figure*}

 From the Table.~\ref{t2} we can discover that all the models with SA-GD training algorithm achieved better performance than these utilized traditional gradient descent algorithm, in spite of the different learning rate strategy or whether considering the learning rate into the transition probability functions. These experimental results validate that the gradient descent algorithm improved with the thought of simulated annealing can indeed enhance the generalization ability of the machine learning models. To be specific, methods SA-Fixed and SA-Fixed-LR which endowing the models the ability of mounting hills surpass the traditional gradient descent by 0.73\% and 0.88\% respectively on the benchmark dataset CIFAR-10. Similarly, the methods SA-Exp and DSA-Exp which utilize the exponentially decay learning rate strategy bring an enhancement of 0.63\% and 0.71\% respectively to the traditional method. Moreover, on the more challenging dataset CIFAR-100, the enhancements of test accuracy that SA-GD methods brings to the traditional gradient descent all surpass 0.8\%, which can be regarded to be significant. 
 
 Besides, the method DSA-Exp which takes exponentially decay learning rate into consideration and employs elaborated transition probability function achieves the highest mean test accuracy on every dataset. However, when comparing it to the SA-Exp method without considering the learning rate, the improvement that DSA-Exp method brings is slightly enhanced. In addition, the extent of enhancement is slightly lower compared to these of SA methods employing fixed learning rate on dataset CIFAR-10. As a result, how to designing the more proper transition probability functions with or without taking learning rate into consideration will be the future work.

\subsection{Algorithm Analysis}

\textbf{Analysis of stability}

To explore the stability of SA-GD and traditional gradient descent algorithms, we tested methods GD-Exp, SA-Exp, and DSA-Exp which performed better than methods utilizing fixed learning rate for one hundred times respectively. The test accuracy of each experiment obtained is exhibited in the Fig.~\ref{p9}. The mean values of GD-Exp, SA-Exp, and DSA-Exp methods calculated are 89.54\%, 90.17\% and 90.21\% respectively as exhibited in Table.~\ref{t2}. The variance of method GD-Exp calculated is 0.00308\%, which is much higher than 0.00072\% obtained by method SA-Exp and 0.00176\% obtained by method DSA-Exp. This convincingly illustrates that the stability of SA-GD algorithm is better than traditional gradient descent training algorithm.

\textbf{Analysis of Convergence}

To explore the convergence speeds of SA-GD and traditional gradient descent algorithms, we analysis the training and test loss curves of methods GD-Exp, SA-Exp and DSA-Exp. Fig.~\ref{p10} shows the training and test loss curves of these selected methods. It is obviously that the  convergence speeds of different methods are almost the same. Though SA-GD algorithm brings some fluctuations in the first half part of the training process, utilizing SA-GD algorithm does not exert much difference to the speed of convergence compared to the traditional gradient descent algorithm.

\section{Ensemble strategy with SA-GD}

Ensemble learning is one widely-utilized machine learning strategy which aggregates multiple different models to obtain better generalization ability than a single model~\cite{sagi2018ensemble}. It is required that the aggregated models to not only possess relatively good performance respectively, but also diverse to each other ~\cite{polikar2012ensemble}. Snapshot ensembling ~\cite{huang2017snapshot} is one ensemble strategy based on the cyclic cosine annealing learning rate strategy, which aggregates models converging to different local minimums during the training process. Since SA-GD equips the model the ability of mounting hills, the models obtained in different training phases are supposed to converge to different local minimums as well. As a consequence, the models in different phases are diversity but all possess fine performance. Therefore, similar to snapshot ensembling, we can aggregate the models in different training epochs during the training process of SA-GD to enhance the generalization ability of the machine learning models further.

We tested SA-GD ensembling on two challenging benchmark datasets, namely, CIFAR-10 and CIFAR-100. The total number of training epochs is $2000$. The ensembling models are selected every the same interval of epochs since the $800$-th epoch. To strengthen the hill-mounting ability of the cost function further and enhance the probability of the ensembling models converging to various local minimums, we tested some larger expansion factor $\sigma$ which has been stated in Eq.~\ref{eq9}. The experimental configurations, results obtained, and the differences to the baseline are exhibited in the Table.~\ref{t3}. In Table.~\ref{t3}, the column \emph{Ens.} represents whether the certain method utilizes ensembling, the column \emph{Num.} stands for the number of models aggregated, and $\sigma$ is the expansion factor chosen.    

As we can see in the Table.~\ref{t3}, aggregating models during the training process of utilizing traditional gradient descent fails to bring improvement to the final prediction performance on both benchmark datasets. Otherwise, increasing the number of ensembling models will not ameliorate the phenomenon. On the contrary, all methods with SA-GD ensembling strategy can bring astonishing improvement to the final prediction accuracy. More specifically, SA-GD ensembling with $24$ models and expansion factor of $4$ surpasses the baseline by $3.47\%$ on CIFAR-10 dataset and $5.92\%$ on CIFAR-100 dataset. In addition, aggregating $24$ models outperforms aggregating $12$ models, indicating that the diversity of these ensembling models is significant resulted by the SA-GD ensembling strategy. Moreover, utilizing larger expansion factors brings higher enhancement than utilizing smaller expansion factors. This phenomenon validates the statement that larger expansion factor can actually strengthen the hill-mounting ability of the cost function further and make the ensembling models more likely to converge to various local minimums.

\begin{table}[ht]
   \caption{Experimental results of methods with or without SA-GD ensembling.}
    \label{t3}
   \centering
   \scalebox{0.95}{
    \begin{tabular}{|c|c|c|c|c|c|}
    \hline
    Method & Ens. & Num. & $\sigma$ &CIFAR-10 & CIFAR-100  \\
    \hline
    GD & False & - & - & 88.83\% (+0.00\%) & 65.22\% (+0.00\%)\\
    \hline
    GD & True & 12 & - & 88.90\% (+0.07\%) & 65.26\% (+0.04\%)\\
    \hline
    SA-GD & True & 12 & 1 & 89.66\% (+0.83\%) & 66.87\% (+1.65\%) \\
    \hline
    SA-GD & True & 12 & 2 & 89.85\% (+1.02\%) &  67.16\% (+1.94\%)\\
    \hline
    SA-GD & True & 12 & 3 & 90.78\% (+1.95\%)&  68.06\% (+2.84\%)\\
    \hline
    SA-GD & True & 12 & 4 & 91.28\% (+2.45\%) &  69.17\% (+3.95\%)\\
    \hline
    GD & True & 24 & - & 88.80\% (-0.03\%) & 65.30\% (+0.08\%)\\
    \hline
    SA-GD & True & 24 & 1 & 90.11\% (+1.28\%) & 67.53\% (+2.31\%) \\
    \hline
    SA-GD & True & 24 & 2 & 90.97\% (+2.14\%) & 68.99\% (+3.77\%) \\
    \hline
    SA-GD & True & 24 & 3 & 91.53\% (+2.70\%) & 69.53\% (+4.31\%)\\
    \hline
    SA-GD & True & 24 & 4 & \textbf{92.31\% (+3.47\%)} & \textbf{71.14\% (+5.92\%)}\\
    \hline
    \end{tabular}
    }
 \end{table}

\section{Conclusion}
In this paper, we introduced a novel training algorithm named as SA-GD, which adopts the thought of simulated annealing algorithm to improve the performance of traditional gradient descent algorithm. In addtion, we exhibit the evolution and reasoning procedure of SA-GD algorithm step by step.

SA-GD algorithm is supposed to make the loss function leave the areas of local minimums or saddle points, enhance the generalization ability and robustness of models as a result. Experimental results validate that SA-GD can actually improve the model test accuracy significantly without sacrificing speed or stability of model convergence. Moreover, SA-GD algorithm can be utilized as an ensemble learning method and enhance prediction accuracy further. 

We think that the formation of gradient ascent probability function utilized in SA-GD deserves further exploring.


{\small
\begin{thebibliography}{10}

\bibitem{beichl2000metropolis}
Isabel Beichl and Francis Sullivan.
\newblock The metropolis algorithm.
\newblock {\em Computing in Science \& Engineering}, 2(1):65--69, 2000.

\bibitem{bhatt2021continuous}
Shobha Bhatt, Anurag Jain, and Amita Dev.
\newblock Continuous speech recognition technologies—a review.
\newblock {\em Recent Developments in Acoustics}, pages 85--94, 2021.

\bibitem{cai2021study}
Zhicheng Cai and Chenglei Peng.
\newblock A study on training fine-tuning of convolutional neural networks.
\newblock In {\em 2021 13th International Conference on Knowledge and Smart
  Technology (KST)}, pages 84--89. IEEE, 2021.

\bibitem{du2016simulated}
Ke-Lin Du and MNS Swamy.
\newblock Simulated annealing.
\newblock In {\em Search and optimization by metaheuristics}, pages 29--36.
  Springer, 2016.

\bibitem{2010Understanding}
Xavier Glorot and Yoshua Bengio.
\newblock Understanding the difficulty of training deep feedforward neural
  networks.
\newblock {\em Journal of Machine Learning Research}, 9:249--256, 2010.

\bibitem{goodfellow2016deep}
Ian Goodfellow, Yoshua Bengio, and Aaron Courville.
\newblock {\em Deep learning}.
\newblock MIT press, 2016.

\bibitem{gross2012linear}
J{\"u}rgen Gro{\ss}.
\newblock {\em Linear regression}, volume 175.
\newblock Springer Science \& Business Media, 2012.

\bibitem{he2016deep}
Kaiming He, Xiangyu Zhang, Shaoqing Ren, and Jian Sun.
\newblock Deep residual learning for image recognition.
\newblock In {\em Proceedings of the IEEE conference on computer vision and
  pattern recognition}, pages 770--778, 2016.

\bibitem{hinton2007boltzmann}
Geoffrey~E Hinton.
\newblock Boltzmann machine.
\newblock {\em Scholarpedia}, 2(5):1668, 2007.

\bibitem{huang2017snapshot}
Gao Huang, Yixuan Li, Geoff Pleiss, Zhuang Liu, John~E Hopcroft, and Kilian~Q
  Weinberger.
\newblock Snapshot ensembles: Train 1, get m for free.
\newblock {\em arXiv preprint arXiv:1704.00109}, 2017.

\bibitem{ioffe2015batch}
Sergey Ioffe and Christian Szegedy.
\newblock Batch normalization: Accelerating deep network training by reducing
  internal covariate shift.
\newblock In {\em International conference on machine learning}, pages
  448--456. PMLR, 2015.

\bibitem{ketkar2017stochastic}
Nikhil Ketkar.
\newblock Stochastic gradient descent.
\newblock In {\em Deep learning with Python}, pages 113--132. Springer, 2017.

\bibitem{khirirat2017mini}
Sarit Khirirat, Hamid~Reza Feyzmahdavian, and Mikael Johansson.
\newblock Mini-batch gradient descent: Faster convergence under data sparsity.
\newblock In {\em 2017 IEEE 56th Annual Conference on Decision and Control
  (CDC)}, pages 2880--2887. IEEE, 2017.

\bibitem{kingma2014adam}
Diederik~P Kingma and Jimmy Ba.
\newblock Adam: A method for stochastic optimization.
\newblock {\em arXiv preprint arXiv:1412.6980}, 2014.

\bibitem{krizhevsky2012imagenet}
Alex Krizhevsky, Ilya Sutskever, and Geoffrey~E Hinton.
\newblock Imagenet classification with deep convolutional neural networks.
\newblock In {\em Advances in neural information processing systems}, pages
  1097--1105, 2012.

\bibitem{lecun2015deep}
Yann LeCun, Yoshua Bengio, and Geoffrey Hinton.
\newblock Deep learning.
\newblock {\em nature}, 521(7553):436--444, 2015.

\bibitem{li2012brief}
Jing Li, Ji-hang Cheng, Jing-yuan Shi, and Fei Huang.
\newblock Brief introduction of back propagation (bp) neural network algorithm
  and its improvement.
\newblock In {\em Advances in computer science and information engineering},
  pages 553--558. Springer, 2012.

\bibitem{lin2019nesterov}
Jiadong Lin, Chuanbiao Song, Kun He, Liwei Wang, and John~E Hopcroft.
\newblock Nesterov accelerated gradient and scale invariance for adversarial
  attacks.
\newblock {\em arXiv preprint arXiv:1908.06281}, 2019.

\bibitem{lydia2019adagrad}
Agnes Lydia and Sagayaraj Francis.
\newblock Adagrad—an optimizer for stochastic gradient descent.
\newblock {\em Int. J. Inf. Comput. Sci}, 6(5), 2019.

\bibitem{polikar2012ensemble}
Robi Polikar.
\newblock Ensemble learning.
\newblock In {\em Ensemble machine learning}, pages 1--34. Springer, 2012.

\bibitem{qian1999momentum}
Ning Qian.
\newblock On the momentum term in gradient descent learning algorithms.
\newblock {\em Neural networks}, 12(1):145--151, 1999.

\bibitem{ruder2016overview}
Sebastian Ruder.
\newblock An overview of gradient descent optimization algorithms.
\newblock {\em arXiv preprint arXiv:1609.04747}, 2016.

\bibitem{sagi2018ensemble}
Omer Sagi and Lior Rokach.
\newblock Ensemble learning: A survey.
\newblock {\em Wiley Interdisciplinary Reviews: Data Mining and Knowledge
  Discovery}, 8(4):e1249, 2018.

\bibitem{sexton1999beyond}
Randall~S Sexton, Robert~E Dorsey, and John~D Johnson.
\newblock Beyond backpropagation: using simulated annealing for training neural
  networks.
\newblock {\em Journal of Organizational and End User Computing (JOEUC)},
  11(3):3--10, 1999.

\bibitem{shi2020rmsprop}
Naichen Shi, Dawei Li, Mingyi Hong, and Ruoyu Sun.
\newblock Rmsprop converges with proper hyper-parameter.
\newblock In {\em International Conference on Learning Representations}, 2020.

\bibitem{simonyan2014very}
Karen Simonyan and Andrew Zisserman.
\newblock Very deep convolutional networks for large-scale image recognition.
\newblock {\em arXiv preprint arXiv:1409.1556}, 2014.

\bibitem{srivastava2014dropout}
Nitish Srivastava, Geoffrey Hinton, Alex Krizhevsky, Ilya Sutskever, and Ruslan
  Salakhutdinov.
\newblock Dropout: a simple way to prevent neural networks from overfitting.
\newblock {\em The journal of machine learning research}, 15(1):1929--1958,
  2014.

\bibitem{tolstikhin2021mlp}
Ilya Tolstikhin, Neil Houlsby, Alexander Kolesnikov, Lucas Beyer, Xiaohua Zhai,
  Thomas Unterthiner, Jessica Yung, Daniel Keysers, Jakob Uszkoreit, Mario
  Lucic, et~al.
\newblock Mlp-mixer: An all-mlp architecture for vision.
\newblock {\em arXiv preprint arXiv:2105.01601}, 2021.

\bibitem{van1987simulated}
Peter~JM Van~Laarhoven and Emile~HL Aarts.
\newblock Simulated annealing.
\newblock In {\em Simulated annealing: Theory and applications}, pages 7--15.
  Springer, 1987.

\bibitem{vaswani2017attention}
Ashish Vaswani, Noam Shazeer, Niki Parmar, Jakob Uszkoreit, Llion Jones,
  Aidan~N Gomez, {\L}ukasz Kaiser, and Illia Polosukhin.
\newblock Attention is all you need.
\newblock In {\em Advances in neural information processing systems}, pages
  5998--6008, 2017.

\bibitem{voulodimos2018deep}
Athanasios Voulodimos, Nikolaos Doulamis, Anastasios Doulamis, and Eftychios
  Protopapadakis.
\newblock Deep learning for computer vision: A brief review.
\newblock {\em Computational intelligence and neuroscience}, 2018, 2018.

\bibitem{wiering2012reinforcement}
Marco~A Wiering and Martijn Van~Otterlo.
\newblock Reinforcement learning.
\newblock {\em Adaptation, learning, and optimization}, 12(3), 2012.

\bibitem{wolf2020transformers}
Thomas Wolf, Julien Chaumond, Lysandre Debut, Victor Sanh, Clement Delangue,
  Anthony Moi, Pierric Cistac, Morgan Funtowicz, Joe Davison, Sam Shleifer,
  et~al.
\newblock Transformers: State-of-the-art natural language processing.
\newblock In {\em Proceedings of the 2020 Conference on Empirical Methods in
  Natural Language Processing: System Demonstrations}, pages 38--45, 2020.

\bibitem{zeiler2012adadelta}
Matthew~D Zeiler.
\newblock Adadelta: an adaptive learning rate method.
\newblock {\em arXiv preprint arXiv:1212.5701}, 2012.

\end{thebibliography}

}
\end{document}